%% file: script.tex
\documentclass{article}

\usepackage[verbose=true,letterpaper]{geometry}
\AtBeginDocument{
  \newgeometry{
    textheight=9in,
    textwidth=5.5in,
    top=1in,
    headheight=12pt,
    headsep=25pt,
    footskip=30pt
  }
}

\renewenvironment{abstract}%
{%
  \vskip 0.075in%
  \centerline%
  {\large\bf Abstract}%
  \vspace{0.5ex}%
  \begin{quote}%
}
{
  \par%
  \end{quote}%
  \vskip 1ex%
}

\makeatletter
\providecommand{\maketitle}{}
\renewcommand{\maketitle}{%
  \par
  \begingroup
    \renewcommand{\thefootnote}{\fnsymbol{footnote}}
    \renewcommand{\@makefnmark}{\hbox to \z@{$^{\@thefnmark}$\hss}}
    \long\def\@makefntext##1{%
      \parindent 1em\noindent
      \hbox to 1.8em{\hss $\m@th ^{\@thefnmark}$}##1
    }
    \thispagestyle{empty}
    \@maketitle
    \@thanks
  \endgroup
  \let\maketitle\relax
  \let\thanks\relax
}

\newcommand{\@toptitlebar}{
  \hrule height 4\p@
  \vskip 0.25in
  \vskip -\parskip%
}
\newcommand{\@bottomtitlebar}{
  \vskip 0.29in
  \vskip -\parskip
  \hrule height 1\p@
  \vskip 0.09in%
}

\providecommand{\@maketitle}{}
\renewcommand{\@maketitle}{%
  \vbox{%
    \hsize\textwidth
    \linewidth\hsize
    \vskip 0.1in
    \@toptitlebar
    \centering
    {\LARGE\bf \@title\par}
    \@bottomtitlebar
      \def\And{%
        \end{tabular}\hfil\linebreak[0]\hfil%
        \begin{tabular}[t]{c}\bf\rule{\z@}{24\p@}\ignorespaces%
      }
      \def\AND{%
        \end{tabular}\hfil\linebreak[4]\hfil%
        \begin{tabular}[t]{c}\bf\rule{\z@}{24\p@}\ignorespaces%
      }
      \begin{tabular}[t]{c}\bf\rule{\z@}{24\p@}\@author\end{tabular}%
      
    \vskip 0.3in \@minus 0.1in
  }
}
\makeatother

\usepackage[utf8]{inputenc} 
\usepackage[T1]{fontenc}    
\usepackage{hyperref}       
\usepackage{url}            
\usepackage{booktabs}       
\usepackage{amsfonts}       
\usepackage{nicefrac}       
\usepackage{microtype}      
\usepackage{xcolor}         

\usepackage{adjustbox}
\usepackage{amsfonts} 
\usepackage{amsmath}
\usepackage{bm}
\usepackage{enumitem}
\usepackage{dirtytalk}
\usepackage{multirow}
\usepackage{MnSymbol}
\usepackage{subcaption}
\usepackage{xspace}

\newcommand{\softmax}{\ensuremath{\mathrm{softmax}}}

\newcommand{\qqmn}{FocusLearn\xspace}
\newcommand{\qqmnlong}{Attention Modular Networks\xspace}

\newcommand{\rnn}{RNN\xspace}
\newcommand{\afs}{AFS\xspace}
\newcommand{\afslong}{Attention-based Feature Selection}
\newcommand{\mods}{MNN\xspace}
\newcommand{\anb}{ANB\xspace}
\newcommand{\anblong}{Attention-based Node Bootstrapping}

\newcommand{\ie}{\emph{i.e.,}\xspace}
\newcommand{\eg}{\emph{e.g.,}\xspace}

\newcommand{\etal}{\emph{et al.\@}\xspace}

\newcommand{\myxfig}[1]{\small\textcircled{\scriptsize #1}\xspace}
\newcommand{\myfigone}{\myxfig{1}}
\newcommand{\myfigtwo}{\myxfig{2}}
\newcommand{\myfigthree}{\myxfig{3}}
\newcommand{\myfigfour}{\myxfig{4}}
\newcommand{\myfigfive}{\myxfig{5}}

\newcommand{\myx}[1]{\normalsize\textcircled{\small #1}\xspace}
\newcommand{\myone}{\myx{1}}

\newcommand{\mythree}{\myx{3}}
\newcommand{\myfour}{\myx{4}}

\title{FocusLearn: Fully-Interpretable, High-Performance Modular Neural Networks for Time Series}

\author{%
  Qiqi Su\thanks{Correspondence to: Qiqi Su <qiqi.su@city.ac.uk>}\\
  Department of Computer Science\\
  City, University of London\\
  London, UK \\
  \And
  Christos Kloukinas \\
  Department of Computer Science \\
  City, University of London \\
  London, UK\\
  \AND
  Artur d'Avila Garcez \\
  Department of Computer Science \\
  City, University of London \\
  London, UK\\
}

\begin{document}

\maketitle

\begin{abstract}
Multivariate time series have had many applications in areas from healthcare and finance to meteorology and life sciences. Although deep neural networks have shown excellent predictive performance for time series, they have been criticised for being non-interpretable.
Neural Additive Models, however, are known to be fully-interpretable by construction, but may achieve far lower predictive performance than deep networks when applied to time series.
This paper introduces \qqmn , a fully-interpretable modular neural network capable of matching or surpassing the predictive performance of deep networks trained on multivariate time series. In \qqmn , a recurrent neural network learns the temporal dependencies in the data, while a multi-headed attention layer learns to weight selected features while also suppressing redundant features. Modular neural networks are then trained in parallel and independently, one for each selected feature. This modular approach allows the user to inspect how features influence outcomes in the exact same way as with additive models. Experimental results show that this new approach outperforms additive models in both regression and classification of time series tasks, achieving predictive performance that is comparable to state-of-the-art, non-interpretable deep networks applied to time series.
\end{abstract}

\section{Introduction}

Deep Neural Networks (DNNs) are a popular method for analysing multivariate time series. DNNs can identify hidden patterns in data to obtain accurate approximations of time series \cite{Gao2019TimeSeries}. Unlike classical statistical methods, DNNs make no assumptions about the statistical distribution of the underlying time series \cite{Kaushik2020TimeSeries}. For this reason, they have been shown to improve on more traditional linear time series models such as Support Vector Machines and auto-regressive models \cite{Zhang2003Hybrid}. Transformer-based models and attention mechanisms, originally introduced for natural language processing, have recently been shown to achieve even higher accuracy at forecasting time series \cite{Choi2019XAI, Ge2018XAI} due to their ability to capture long-term dependencies in data.

Despite DNNs' success, there is now widespread recognition that just effective predictive performance is insufficient; DNNs need to be interpretable or explainable, especially in safety-critical applications. As a result, the field of eXplainable Artificial Intelligence (XAI) has gained traction in recent years. XAI seeks to make DNNs more transparent and understandable to humans, and to provide greater trust in AI decisions. Interpretability approaches analyse the model’s weights and features when determining a given output. Explainability seeks to derive meaning from the input data and model outputs to map out the behaviour of the model. XAI has been applied to help developers improve model learning, to help users make more informed decisions, and to identify biases in the trained models \cite{Ngan2023NeuralSymbolic,Wagner2021,White2023Counterfactural}.

XAI research can be divided into post-hoc and ante-hoc. Post-hoc methods aim to explain a trained model. Ante-hoc methods seek to incorporate interpretability into model training. Post-hoc methods can only approximate the underlying model, which can make the explanations less faithful. Ante-hoc approaches may be less flexible as they prescribe a specific architecture or training regime. In general, a trade-off exists between accuracy and interpretability in ante-hoc methods. It is important to develop ante-hoc interpretable DNNs that can outperform traditional methods deemed to be inherently-interpretable, such as decision trees, while matching the performance of non-interpretable approaches \cite{Dubey2022SPAM}. 

To achieve an interpretable DNN for time series, we draw inspiration from modular networks \cite{Caelli1999Modularity} and additive models \cite{Agarwal2021NAM}. Our approach builds on the evidence pointing to the interpretability of linear directions in activation space, particularly linear combinations of neuron activations \cite{Bricken2023Monosemanticity}. Our proposed architecture learns a linear combination of modular networks, each attending to a single selected feature. This process aims to decompose the complexity of DNNs into fundamental units of features, as done by Neural Additive Models (NAMs) \cite{Agarwal2021NAM}. Differently from \cite{Bricken2023Monosemanticity}, we emphasise the relevance of modularity during the learning process, highlighting its significance in steering the model away from convoluted structures that may hinder interpretability. Leveraging the NAM approach, we aggregate the contributions of individual neural networks by summing them up. The incorporation of modularity at learning time in our methodology will be shown to enhance interpretability compared to DNNs. It will be shown that learned modularity allows one to analyse and systematically discern the contributions of each feature, facilitating a more transparent understanding of the DNN's decision-making.

Following the neural additive approach, we propose \qqmn, which is able to handle both time series forecasting and classification tasks for both large and small data sets. \qqmn provides feature importance and the exact depiction of how the model arrives at a certain outcome, offering interpretability at the same level as NAMs. In addition, \qqmn outperforms state-of-the-art interpretable methods, such as NAM itself, and it matches the performance of classic non-interpretable models on all four domains considered in this paper. \qqmn achieves these results by introducing two novel ideas: \emph{(i)} \emph{it uses an attention mechanism for learning the relevant features}; and \emph{(ii) it reuses the attention weights when training each modular network on the final prediction}, as detailed in what follows. 

The paper is organised as follows. Section II discusses Related Work. Section III introduces \qqmn. Section IV contains the main experimental results. Section V elaborates on and discusses variations of the experiments, and Section VI concludes the paper and discusses directions for future work.

\section{Related Work}

Popular XAI methods \textit{Local Interpretable Model-agnostic Explanation} (LIME) \cite{Ribeiro2016LIME} and \textit{SHapley Additive exPlanation} (SHAP) \cite{Lundberg2017SHAP} have been introduced in the hope of overcoming the accuracy-interpretability trade-off in DNNs. However, these methods suffer from low fidelity \cite{Adadi2018XAI,White2020Counterfactual}, instability \cite{Ghorbani2019IML}, and even inaccuracy \cite{Lipton2018IML, Rudin2019XAI} as they seek to approximate the underlying model. 

Many ante-hoc methods have been proposed for time series analysis, including Shaplets \cite{Fang2018XAI,Ye2019Shapelets}, Symbolic Aggregate Approximation \cite{Senin2013XAI}, and Fuzzy Logic-based XAI \cite{Paiva2004XAI,Wang2017XAI}. Often, these methods address either regression or classification tasks, but not both. More importantly, these methods do not seem to scale well with large data sets. 

\subsection{Interpretable Additive Models} 

Adding interpretability to Generalised Additive Models (GAM) \cite{Hastie1986GAM} has been shown applicable to both time series regression and classification problems. GAM variants such as NAM \cite{Agarwal2021NAM}, GA$^2$M \cite{Lou2013GA2M}, Neural Interaction Transparency \cite{Tsang2018NIT} and Neural Basis Model \cite{Radenovic2022NBM}, are computationally efficient and can handle large data sets. For instance, NAM~\cite{Agarwal2021NAM}, an interpretable variant of GAM, learns a linear combination of a family of interpretable neural networks, each assigned to learn a single input feature. This family of networks is trained jointly using backpropagation and it can learn arbitrarily complex shape functions. The impact of a feature on the prediction can then be interpreted by plotting its corresponding shape function along with the coefficients of the linear combination. As such, this plot is the exact depiction of how NAM arrives at a prediction, under the assumption that the pre-defined separation of features holds. NAM is also capable of learning non-linear relationships, making it an attractive choice for many applications. Unfortunately, the predictive performance of these models is no match for more popular, non-interpretable time series models, so that they cannot be regarded as a substitute for non-interpretable models \cite{Dubey2022SPAM}. 

The \textit{Scalable Polynomial Additive Model} (SPAM) \cite{Dubey2022SPAM} builds on NAM by allowing higher-order interactions between features. SPAM is intended to replace non-interpretable models for large scale data due to its higher performance than NAMs. The main difference between SPAM and NAM is that SPAM uses tensor rank decomposition of polynomials such that higher-order feature interactions can be learned. However, the explanations provided by SPAM cannot be readily visualised in the same way as NAM and, as a result, they are harder to interpret than NAM. 

\subsection{Other Relevant Topics}

\subsubsection{DNN for Multivariate Time Series Forecasting}

Recurrent neural networks (\rnn) \cite{Elman1990RNN} are very popular for analysing multivariate time series as they remember past inputs to decide on future outcomes. Many variants of \rnn have been proposed over the years such as Long Short-Term Memory (LSTM) \cite{Hochreiter1997LSTM}, Bidirectional LSTM (BLSTM) \cite{Schuster1997BLSTM}, and Gated Recurrent Units (GRU) \cite{Cho2014GRU}. LSTM and GRU were introduced to overcome the gradient vanishing problem faced by \rnn{}s \cite{Bengio1994RNN}, and in turn have been shown to be useful for learning long-term dependencies \cite{Chung2014LSTM}. Since LSTM has a more complex architecture than GRU, training a GRU requires fewer computational resources. On the other hand, LSTM can handle large amounts of data with more parameters to optimise performance. LSTM is a very popular choice for modelling multivariate time series problems across a variety of applications \cite{Huang2019RNN, Sak2014LSTM, Sunny2020LSTM}, although they are considered to be non-interpretable. 

\subsubsection{Attention Mechanism for Feature Selection}

Attention mechanisms have been used as a way to provide interpretability by offering insight into the most relevant features. Temporal Fusion Transformer (TFT) \cite{Lim2021TFT} was introduced for a multi-horizon forecasting task with added interpretability by visualising the persistent temporal pattern using attention. A limitation of TFT is that it requires a large amount of data to achieve good predictive performance \cite{Zhang2022TFT}. Interpretable Temporal Attention Network (ITANet) \cite{Zhou2022ITANet} used an attention mechanism to infer the importance of government interventions for COVID19 predictions. ITANet only provides feature importance as an explanation. 


\begin{figure*}[!t]
\centerline{\includegraphics[width=.8\textwidth]{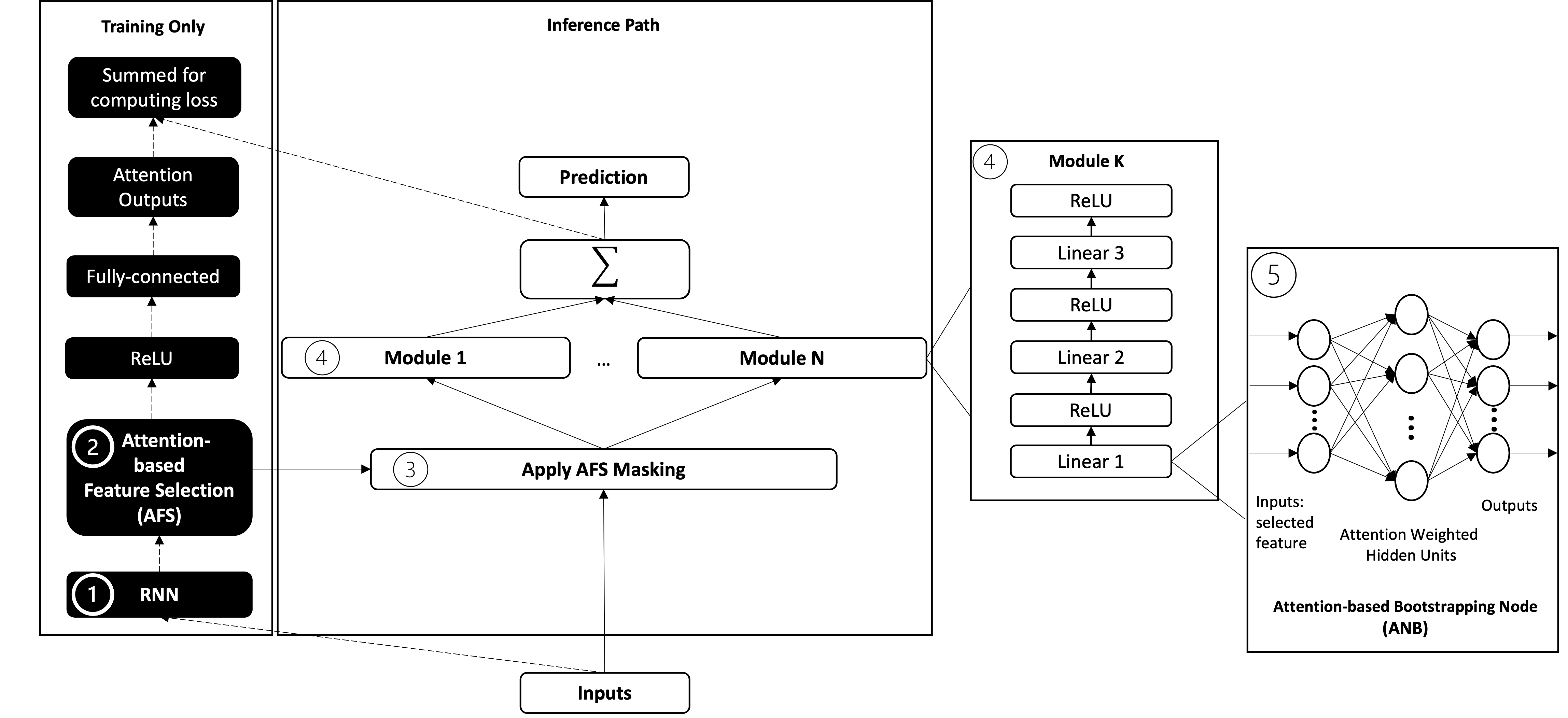}}
\caption{\textbf{\qqmnlong (\qqmn) architecture.} \qqmn consists of two main paths (training and inference) with five main components: \myfigone \textit{Recurrent Neural Network (\rnn)} (see Section~\ref{sec: rnn}), \myfigtwo \textit{\afslong{} (\afs)} with \myfigthree AFS masking (see Section~\ref{sec: afs}), and \myfigfour a group of \textit{interpretable Modular Neural Networks (\mods)} that learn from the top $n$ input features selected by the \afs (see Section~\ref{sec: mod_main}). The \myfigfive \textit{\anblong{} (\anb)} in each \textit{module}'s first layer is weighted by the \textit{\afs}'s attention weights (see Section~\ref{sec: mod_main}). \mods{}s outputs are then aggregated. All components are required for training, but to maintain the interpretability of \qqmn, only the components in the \textit{inference path} box are used after training (see Section~\ref{sec: paths}).} 
\label{fig: architecture}
\end{figure*}

Feature selection has been an effective approach for preparing high-dimensional data for a variety of Machine Learning tasks \cite{Gui2019AFS}. Several different architectures that incorporate attention into DNNs for feature selection have been proposed recently, \eg Attention-based Feature Selection (AFS) \cite{Gui2019AFS}, Multiattention-based Feature Selection (MFS) \cite{Cao2021MFS}, and Attentive Interpretable Tabular Learning (TabNet) \cite{Arik2019TabNet}. Attention mechanisms seek to learn the temporal relationships in the data, especially when used in conjunction with \rnn{}s, better than traditional feature selection methods --- wrapper feature selection methods tend to suffer from high computational complexity, while filter methods evaluate feature weights using only generic characteristics of the data \cite{Cao2021MFS, Gui2019AFS}. In AFS and MFS, several attention mechanisms are utilised in an \textit{attention module} to select features directly from the input. A \textit{learning module} then deploys different DNNs to learn from the selected inputs. 
In both AFS and MFS, interpretability only exists in the \textit{attention module} and not in the learning module, the main part of the system. TabNet introduced a sequential attention mechanism to choose which features to use for each decision step. Although TabNet achieved state-of-the-art results compared to non-interpretable DNNs, the explanations provided by TabNet only concern feature attributions.  
 
\section{\qqmn -- An Attention Modular Network}

\qqmn\footnote{\url{https://github.com/qisuqi/FocusLearn}} is in the category of interpretable additive models, such as NAM and SPAM. Although, like \qqmn, NAM can provide both feature importance and explanation plots for model outcomes, its predictive performance can be poor (see Section IV). SPAM addresses the issue of performance but only offers feature importance explanations. As a result, the goal of this research was to develop a model that performs better than NAM while maintaining NAM's level of explainability. To accomplish this, inspiration was taken from AFS and MFS, where an attention mechanism is used for feature selection. Learning only the most salient features has been shown to be a useful method for improving interpretability and for allocating learning capacity more effectively \cite{Cao2021MFS, Gui2019AFS}. The entirety of the inference path of \qqmn is interpretable since the computation for prediction is simply a series of linear combinations of interpretable networks. Predictive performance is improved, not by making that step nonlinear, but from a better selection of features using attention to learn the modular networks and to ignore features that have no contribution to the model.  

\subsection{Recurrent Neural Network (\rnn)} \label{sec: rnn}

Inputs are first processed by the \rnn component, shown in Fig.~\ref{fig: architecture} \myone , which is responsible for learning the time-dependencies in the input data. The \rnn component used in this paper is an LSTM network, but it can be replaced easily by an alternative architecture such as GRU, BLSTM or Transformer \cite{Katharopoulos2020Trans}. In the hidden layer of an LSTM, there are some special units called memory cells that are recurrently connected, as well as their corresponding gate units, namely input gate, forget gate, and output gate \cite{Hochreiter1997LSTM}. To avoid the problem of vanishing or exploding gradients and unstable hidden representations due to the highly dynamic nature of their state updates, layer normalisation is applied to the \rnn outputs \cite{Ba2016LayerNorm}. It is responsible for re-centering and re-scaling the activation values to obtain zero mean and unit variance, and facilitate the propagation of gradients through the network. 

\subsection{\afslong{} (\afs)} \label{sec: afs}

This component of \qqmn is inspired by the Multi-head Attention (MHA) proposed in \cite{Vaswani2017Attention}. Using an attention mechanism for feature importance or feature selection is often criticised, as using the attention weights of the MHA alone cannot be indicative of the importance of a feature, given that different weights assigned to different parts of the input data are used in each head \cite{Lim2021TFT}. Therefore, instead of making each head attend to a sub-sequence of the input, the MHA is modified here so each attention head computes the attention weight of each input feature independently. It is worth noting that this modification does not change the mathematical concept of the original MHA -- it merely changes how the attention is applied.

\begin{equation} \label{eq: ah_main}
    AH_i = \mathrm{Attention}(\bm{Q}\bm{W}^Q_i, \bm{K}\bm{W}^K_i, \bm{V}\bm{W}^V_i)
\end{equation}
\begin{equation} \label{eq: multihead_attn_main}
    \mathrm{MultiHead} (\bm{Q}, \bm{K}, \bm{V}) = [AH_1, AH_2,...,AH_n] \bm{W}_{AH}
\end{equation}

Given the final hidden states of the \rnn component, $H = h_1,h_2,...,h_T$, the attention function in \afs, \eqref{eq: ah_main}, is applied to each $H_i$, where each $AH_i$ is an attention head of the MHA, and $\bm{Q}\bm{W}^Q_i \in \mathbb{R}^{d_q \times d_v}$, $\bm{K}\bm{W}^K_i \in \mathbb{R}^{d_k \times d_v}$, $\bm{V}\bm{W}^V_i \in \mathbb{R}^{d_q \times d_v}$ are the projections of attention weights for $\bm{Q}$, $\bm{K}$, $\bm{V}$, respectively. In \eqref{eq: multihead_attn_main}, $\bm{W}_{AH} \in \mathbb{R}^{(AH \cdot d_v) \times d_q}$ linearly combines concatenated outputs from all attention heads, thus completing the MHA function. Finally, $\bm{W}_{AH}$ is averaged to compute the attention weights for each $H_i$ by applying a feature-wise pooling operation. The output of the \afs component, $\bm{F_i}$, is obtained by applying the \softmax{} function to the averaged attention weights, $\text{Attn}(H_i)$, such that $\bm{F_i} = \softmax(\text{Attn}(H_i))$. $F_i$ allows us to select the top $n$ features from the inputs in component \mythree of \qqmn, \textit{Apply \afs Masking}, Fig.~\ref{fig: architecture} \mythree, filtering out features that do not contribute much to model prediction. In \qqmn, $n$ can be pre-defined or a hyper-parameter. To ensure that gradients are propagated adequately through the \mods (discussed in more detail in Section~\ref{sec: mod_main}), $\bm{F}$ is put through a Rectified Linear Unit (ReLU) and a fully-connected layer so that the loss of the \rnn-\afs components can be calculated (discussed in more detail in Section~\ref{sec: loss_func_main}). 

\subsection{Modular Neural Networks (\mods)} \label{sec: mod_main} 

The idea of modularity has been researched as early as the 1980s and adopted in neural information processing in the late 1990s \cite{Caelli1999Modularity}. It was recognised then that traditional neural networks are black-boxes and \mods{}s were more interpretable alternatives. \mods{}s follow the divide-and-conquer principle \cite{Varela2021Modularity} and are inspired by the important biological fact that neurons in human brains are sparsely connected in a clustered and hierarchical fashion, rather than completely connected \cite{Caelli1999Modularity}. \mods{}s use several smaller simple Neural Networks (NNs), where each NN focuses on a different part of the same problem and their outputs are combined together into the outcome for the entire network. In this way, \mods{}s break down complex learning problems in resolving large-scale tasks.

Each module in the \mods component of \qqmn consists of a single selected input feature trained independently, so that the impact of each feature on the prediction is independent of other features. It is worth noting that the \mods component reads the top $n$ features directly from the inputs. Thus, this component methodologically belong to the GAM family, $g(\mathbb{E}(y)) = \beta + f_1(x_1) + f_2(x_2) + \cdots + f_n(x_n)$, where $y$ is the target variable, $g$ is the link function, $\bm{x} = x_1, x_2, ... , x_n$ is the input (or the selected input in the case of \qqmn), and each $f_i$ is a univariate shape function with $\mathbb{E}(f_i)=0$.

The \mods component renders the \qqmn interpretable because each univariate shape function is parameterised by a NN through a series of linear transformations, as done with NAM. The main difference between \qqmn and NAM is that each feature is initially weighted using the attention weights $\bm{F}$. This will be shown to improve predictive performance considerably compared to NAM. To achieve feature weighting, we introduce \anblong{} (\anb) to each module (see Fig.~\ref{fig: architecture} \myfour ). The objective is to learn the weights based on the inputs factored by the attention weights and shifted by a bias. For each module, Xavier Initialisation \cite{Glorot2010XavierInit} sets the biases (to zeros) and weights at each layer. The previously-computed attention weights for the selected features, $\bm{F}$, are then multiplied by the initialised weights, $\bm{W}_\text{init}$, as in \eqref{eq: anb1}. Finally, the ReLU activation function is applied to each \anb for each scalar input from $x$, as in \eqref{eq: anb2}. In practice, \anb is applied to the first layer of each module (Fig.~\ref{fig: architecture} \myfour , Linear 1) so each $\bm{F_i}$ can be used directly. The modular networks are then trained jointly with the \rnn and \afs components using back-propagation, as follows.

\begin{equation}\label{eq: anb1}
    \bm{W}_{\text{mod}_{i}} = \bm{W}_{\text{init}_i} * \bm{F_i}
\end{equation}
\begin{equation}\label{eq: anb2}
    \anb(x) = \text{ReLU} ((x - b) * \bm{W}_\text{mod})
\end{equation}

\subsection{Learning Rate Scheduler}

Since \qqmn resembles the Transformer architecture, we employ the Cosine Annealing learning rate (LR) warm-up scheduler \cite{Loshchilov2016Cosine} to further accelerate convergence, stabilise the training, and hopefully improve generalisation using the Adam optimisation algorithm \cite{Liu2019LR}. Unlike fixed or decreasing LR strategies, the LR warm-up gradually increases LR from zero to the specified LR in initial training iterations. This is particularly important for adaptive optimisation, as in the case of Adam, which leverage bias correction factors, potentially causing higher variance in early iterations. Additionally, the \rnn component's use of layer normalisation can contribute to elevated gradients in the initial phases.

\subsection{Loss Functions}\label{sec: loss_func_main} 

To ensure that gradients flow through all components of \qqmn and improve convergence during training, the \rnn-\afs and \mods components are trained jointly, with each one aiming to minimise the Mean Squared Error Loss ($\frac{1}{N} \sum (y_i - \Tilde{y_i})^2$) in the case of regression tasks, and a Binary Cross-Entropy with Logits Loss ($- \frac{1}{N} \sum_{i=1}^N (y_i * log(\Tilde{y_i}) + (1 - y_i) * log(1 - \Tilde{y_i}))$) in the case of classification tasks, where $y$ is the target value, $\Tilde{y_i}$ is the predicted value, and $N$ is the number of data points. The \qqmn loss function to minimise is the sum: $\text{Loss}_{\text{\qqmn}} = \text{Loss}_{\text{\rnn-\afs}} + \text{Loss}_{\text{mod}}$.

\subsection{Training and Inference Path} \label{sec: paths}

All components of \qqmn are required in the training phase, but only the inference path is required at deployment time. To ensure the explainability of the deployed model, the inference path includes by construction only the \mods component. This means that at inference time the \rnn-\afs components are excluded from the forward pass. Without gradients traversing through the \rnn-\afs components, the utility of \anb remains unrealised, as the gradients would not be back-propagated to the original inputs. In response to this, we implemented a streamlined inference path where components linked by dashed arrows in the \textit{training only} box of Fig.~\ref{fig: architecture} 
are omitted. This means that the \rnn-\afs components exclusively handle feature selection during training, and once the top $n$ features are selected, \qqmn becomes as explainable as NAM, as it no longer relies on the excluded components for decision-making. 

\section{Experimental Results} \label{sec: exp_results}

\textbf{Predictive performance} was evaluated on two regression and two classification tasks against state-of-the-art interpretable and non-interpretable methods\footnote{Although NAM was not explicitly designed for time series, its flexibility allowed it to be applied to time series data \cite{Jo2023NAM,Vetter2023NAM}.}: 

\begin{itemize}[left=0pt]
  \item \textbf{Neural Additive Model (NAM)} \cite{Agarwal2021NAM}. \qqmn is an extension of NAM with the objective of maintaining interpretability while improving upon its predictive performance; 
  
  \item \textbf{Scalable Polynomial Additive Model (SPAM)} \cite{Dubey2022SPAM}. Although SPAM uses feature interactions to produce more accurate predictions than NAM, SPAM cannot produce NAM-like feature curves as explanations; 
  
  \item \textbf{Long Short-Term Memory (LSTM)} \cite{Hochreiter1997LSTM}. Stand-alone LSTM offers a baseline for evaluating multivariate time series models, but it is non-interpretable; 

  \item \textbf{Extreme Gradient Boosted Trees (XGBoost)}\footnote{\url{https://xgboost.readthedocs.io/en/stable/}}. 
  Although a single tree can be interpretable, interpretability degrades rapidly as the number of trees grows. Both NAM and SPAM are compared with XGBoost in earlier work. 
  
\end{itemize} 

Experimental results are summarised in \tablename~\ref{tab: comparison_results}. The top 10 features are selected for all data sets (\ie we use \mods with 10 modules). \qqmn outperforms both interpretable models NAM and SPAM\footnote{NAM and SPAM implementations lacked an \textit{inference step} for making predictions on unseen data. Results reported here are based on test set performance. We have implemented an inference step for NAM and SPAM in order to make a direct comparison with \qqmn on the test sets.} in both regression and classification tasks. This happened for every standard metric used, namely, Symmetric Mean Absolute Percentage Error (sMAPE), Mean Absolute Scaled Error (MASE) and Weighted Absolute Percentage Error (WAPE) for regression, accuracy, F1 score and Area Under the Curve (AUC) for classification tasks. \qqmn also has significantly faster computation speed than both NAM and SPAM. 

\begin{table}[t]
\begin{center}
\caption{Average test set results of \qqmn compared with other methods on four data sets for time series regression and classification. For regression tasks, lower results are better ($\downarrow$). For classification tasks, higher results are better ($\uparrow$).}
\label{tab: comparison_results}
\vspace*{-\baselineskip}
\begin{adjustbox}{width=\columnwidth}
\begin{tabular}{lllllr}
\hline
\multicolumn{6}{c}{Regression $\downarrow$} \\
\hline
\hline
\\[-1em]
& & SMAPE & MASE & WAPE & Speed (s) \\
\\[-1em]
\hline
\multirow{5}{*}{Air} & \color{blue}{\textbf{\qqmn}} & $\mathbf{0.2640}^{**}$ & ~$\mathbf{0.1654}^{**}$ & $\mathbf{0.1637}^{**}$ & 1,198.10 \\
& \color{blue}{\textbf{NAM}} & $0.5221^*$ & ~$0.4166^*$ & $0.4206^*$ & 3,598.02 \\
& \color{blue}{\textbf{SPAM}} & 0.5231 & ~0.2765 & 0.2778 & 1,571.26 \\
& \color{red}{XGBoost(50)} & $0.2704^{**}$ & ~$0.1704^{**}$ & $0.1706^{**}$ & 15.42 \\
& \color{red}{LSTM} & $0.2935^*$ & ~$0.1774^{**}$ & $0.1772^{**}$ & 1,289.57 \\
\hline
\multirow{5}{*}{OtiReal} & \color{blue}{\textbf{\qqmn}} & $0.2105^* $ & ~$1.4829$ & $0.0806^{**}$ & 96.32 \\
& \color{blue}{\textbf{NAM}} & $0.8543^*$ & $12.3317$ & 0.6661 & 238.95 \\
& \color{blue}{\textbf{SPAM}} & 0.2951 & ~2.3075 & $0.1249^*$ & 119.73 \\
& \color{red}{XGBoost(100)} & $0.9574^{**}$ & ~$\mathbf{1.0840}^{**}$ & $0.6846^{**}$ & 1.29 \\
& \color{red}{LSTM} & $\mathbf{0.1809}^*$ & ~1.2694 & $\mathbf{0.0666}^*$ & 92.77 \\
\hline
\multicolumn{6}{c}{Classification $\uparrow$} \\
\hline
\hline
\\[-1em]
& & Accuracy & F1 & AUC & Speed (s) \\
\\[-1em]
\hline
\multirow{5}{*}{EEG} & \color{blue}{\textbf{\qqmn}} & $\mathbf{0.8203^*}$ & $\mathbf{0.8131^*}$ & $\mathbf{0.8203^*}$ & 24.28\\
& \color{blue}{\textbf{NAM}} & $0.5430^{**}$ & $0.0115^{**}$ & $0.5020^{**}$ & 136.56\\
& \color{blue}{\textbf{SPAM}} & $0.5000^*$ & $0.0159^*$ & $0.5002^*$ & 115.61 \\
& \color{blue}{\textbf{XGBoost(5)}} & $0.6959^{**}$ & $0.6169^*$ & $0.6824^{**}$ & 0.02\\
& \color{red}{LSTM} & $0.5000^{**}$ & 0.2667 & $0.5^{**}$ & 53.53\\
\hline
\multirow{5}{*}{Weather} & \color{blue}{\textbf{\qqmn}} & $\mathbf{0.8518}^{**}$ & $0.5157^*$ & $0.6791^{**}$ & 1,192 \\
& \color{blue}{\textbf{NAM}} & $0.8169^{**}$ & 0.2442 & $0.5687^*$ & 1,552.91 \\
& \color{blue}{\textbf{SPAM}} & $0.8369^{**}$ & 0.4305 & $0.6444^*$ & 1,231.40 \\
& \color{blue}{\textbf{XGBoost(5)}} & $0.8473^*$ & $\mathbf{0.5508}^*$ & $\mathbf{0.7009}^*$ & 20.67\\
& \color{red}{LSTM} & $0.8419^{**}$ & $0.4592^*$ & $0.6509^{**}$ & 664.46 \\

\hline
\end{tabular}
\end{adjustbox}
\end{center}
$^{(*)}$ Variance across experiments is $< 0.05$. \\
\color{black}$^{(**)}$ Variance across experiments is $< 0.01$. \\
\color{blue}{\textbf{FocusLearn, NAM, SPAM, XGBoost(5) are interpretable.} \color{red}{\textbf{XGBoost(50), XGBoost(100), LSTM are non-interpretable.}} \color{black}The number in parentheses after XGBoost is the optimal number of trees; XGBoost is considered to be non-interpretable when the number of trees exceeds 10.} \\
\color{black}\textit{Note 1: Means and standard deviations are reported from 5 random trials with optimal hyper-parameters for each task (see Appendix A 
of \cite{Su2023}).}\\
\textit{Note 2: Data set details are provided in Appendix B 
of \cite{Su2023}. 
\textit{Air} \cite{Zhang2017Air} predicts future particulate matter values (PM$_{2.5}$), 
\textit{OtiReal} \cite{Christensen2021Oticon} is used to predict future hearing aid usage, \textit{EEG} \cite{Roesler2013EEG} predicts eyes-open or closed states, and \textit{Weather} predicts if it will rain the next day.}
\vspace*{-.5\baselineskip}
\end{table}

\begin{figure}[!t] 
\centering 
\subfloat[\qqmn]{\includegraphics[width=\columnwidth, height=2.5cm]{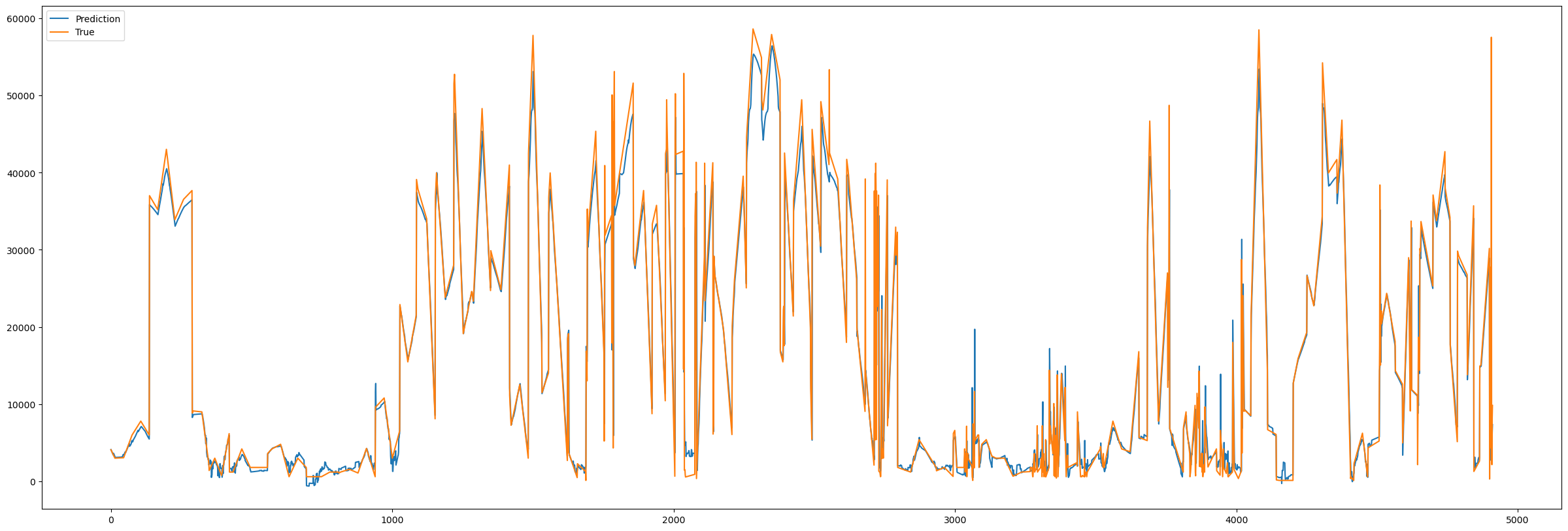}} 
\vfill
\subfloat[LSTM]{\includegraphics[width=\columnwidth, height=2.5cm]{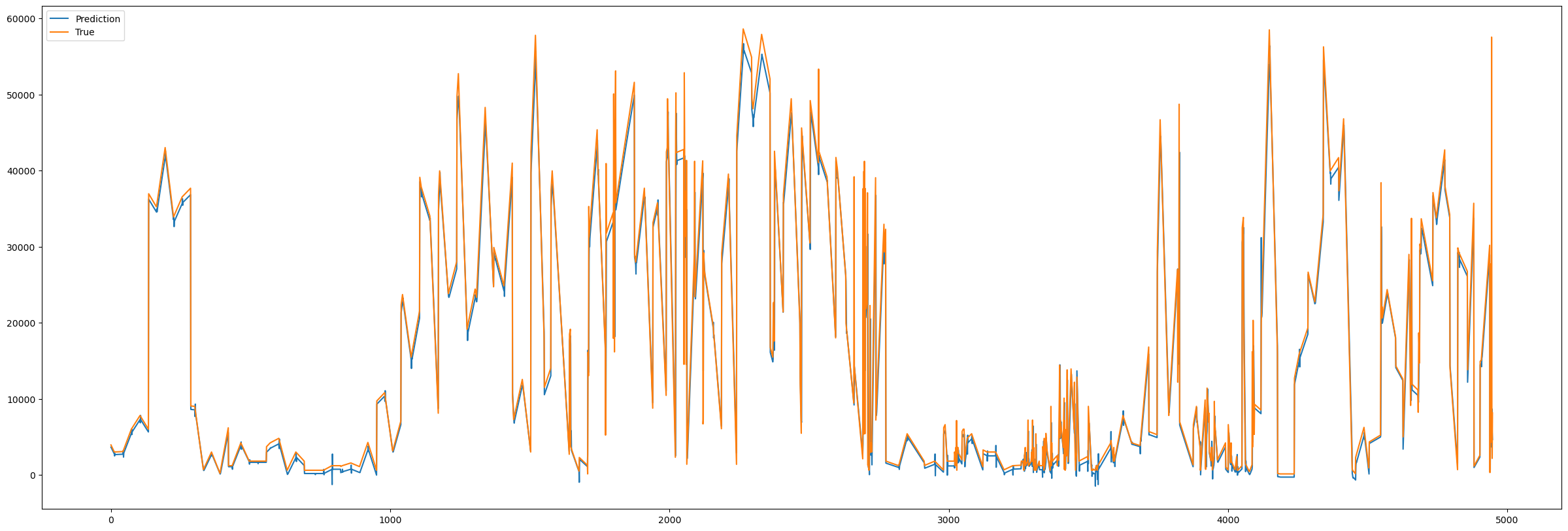}} 
\vfill 
\subfloat[XGBoost]{\includegraphics[width=\columnwidth, height=2.5cm]{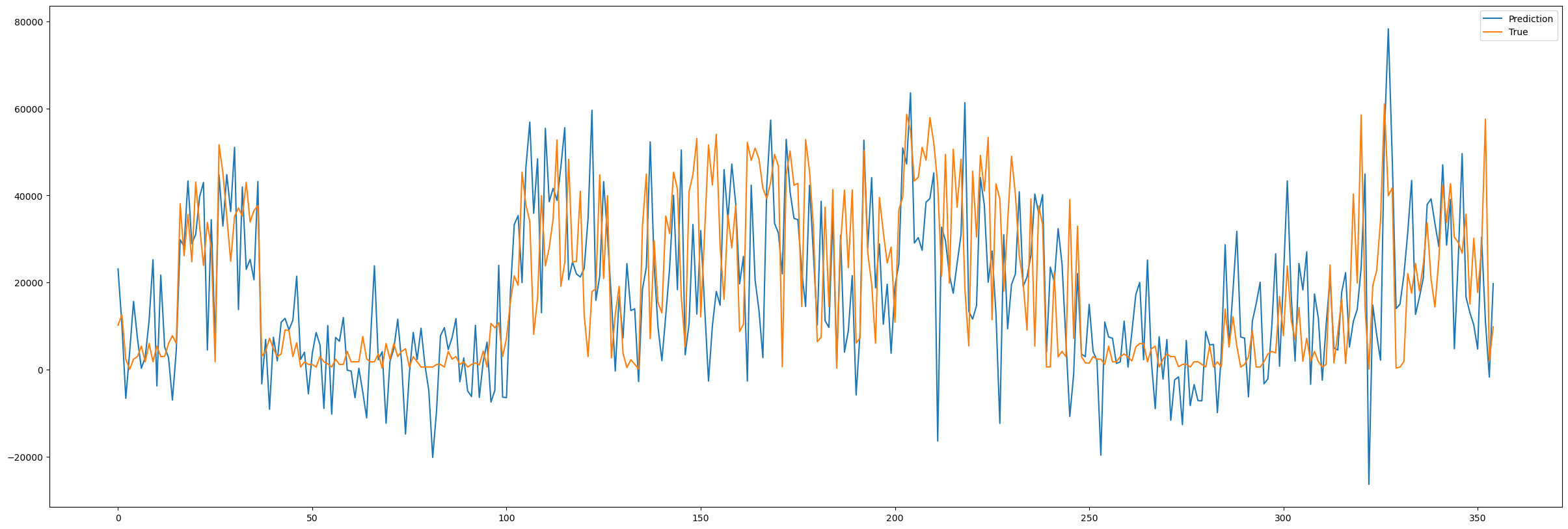}} 
\caption{Prediction visualisations. The x-axis represents predicted time-steps and the y-axis represents feature values. \textit{Note: data are pre-processed differently for XGBoost, which supports missing values by default, whereas missing values are interpolated for training with \qqmn and LSTM.}} 
\label{fig: predvis} 
\end{figure}

As \tablename~\ref{tab: comparison_results} shows, \qqmn also matches the performance of non-interpretable approaches on the \textit{OtiReal} and \textit{Weather} data sets, or it outperforms non-interpretable approaches (\textit{Air} and \textit{EEG} data sets). In the case of the \textit{OtiReal} results, \qqmn and LSTM had the lowest SMAPE and WAPE but did not have a good MASE. Reversely, XGBoost had the lowest MASE error, but its SMAPE and WAPE errors were among the worst. Both SMAPE and WAPE are percentage-based error estimators, whereas MASE evaluates the ability of a model relative to a simple benchmark model. Therefore, given the discrepancy in results depending on the choice of metric, a visual inspection of the prediction results would be helpful to decide which model performed better in the \textit{OtiReal} data set. While metrics are essential for quantitative evaluation, the ultimate goal is to achieve predictions that align with the underlying pattern in the data and expert intuition. Fig.~\ref{fig: predvis} shows that \qqmn and LSTM are better at capturing the pattern and the trend in the \textit{OtiReal} data set than XGBoost: the predicted values are much better aligned with the true values. Fig.~\ref{fig: predvis} also shows that the magnitude of the true values vary significantly in the \textit{OtiReal} data set and both SMAPE and WAPE are better at comparing models in this situation than MASE. 

\begin{figure}[!t]
\centering 
\subfloat[\textbf{\textit{Air} data set}. Showing graphs (shape functions) learned by \qqmn at predicting future \textit{PM$_{2.5}$} value given \textit{Carbon Monoxide} and \textit{Sulfur Dioxide} concentration. The shape function plot for \textit{CO} shows an overall positive correlation (highlighted by the superimposed green line) with large variation. Similarly for \textit{SO2} there is a positive correlation until the sudden drop when \textit{SO2} exceeds 320 $\mu g/mg^3$. Notice how simply sampling the input to obtain such correlations would not be informative or feasible in practice with an increased number of input variables.]{\includegraphics[width=.9\columnwidth]{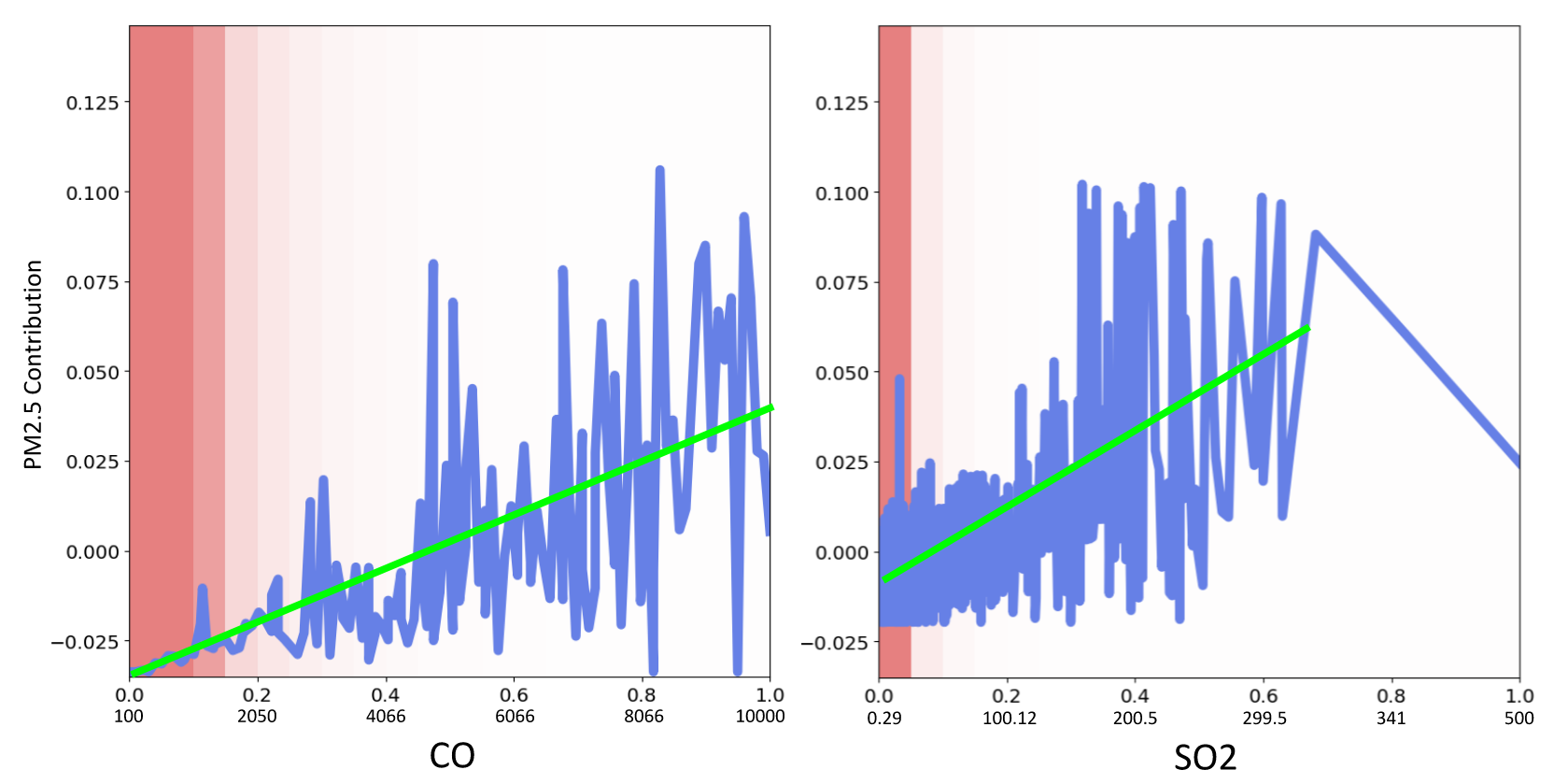} 
\label{fig: air_nam}} 
\vfil
\subfloat[\textbf{\textit{Weather} data set.} The graphs (shape functions) learned by \qqmn predict whether it will rain the following day given \textit{Humidity} and \textit{Wind Direction}. The shape function plot for \textit{Humidity3pm} shows that humidity from 60 to 99\% at 3pm today will likely lead to \textit{rain} tomorrow. The start of the upward trend of the contribution of \textit{humidity} occurs in a region of high data density. The plot for \textit{WindDir3pm} shows higher chances of rain  when there is a northerly wind compared with lower chances of rain when the wind direction changes to a more southerly wind.]{\includegraphics[width=.9\columnwidth]{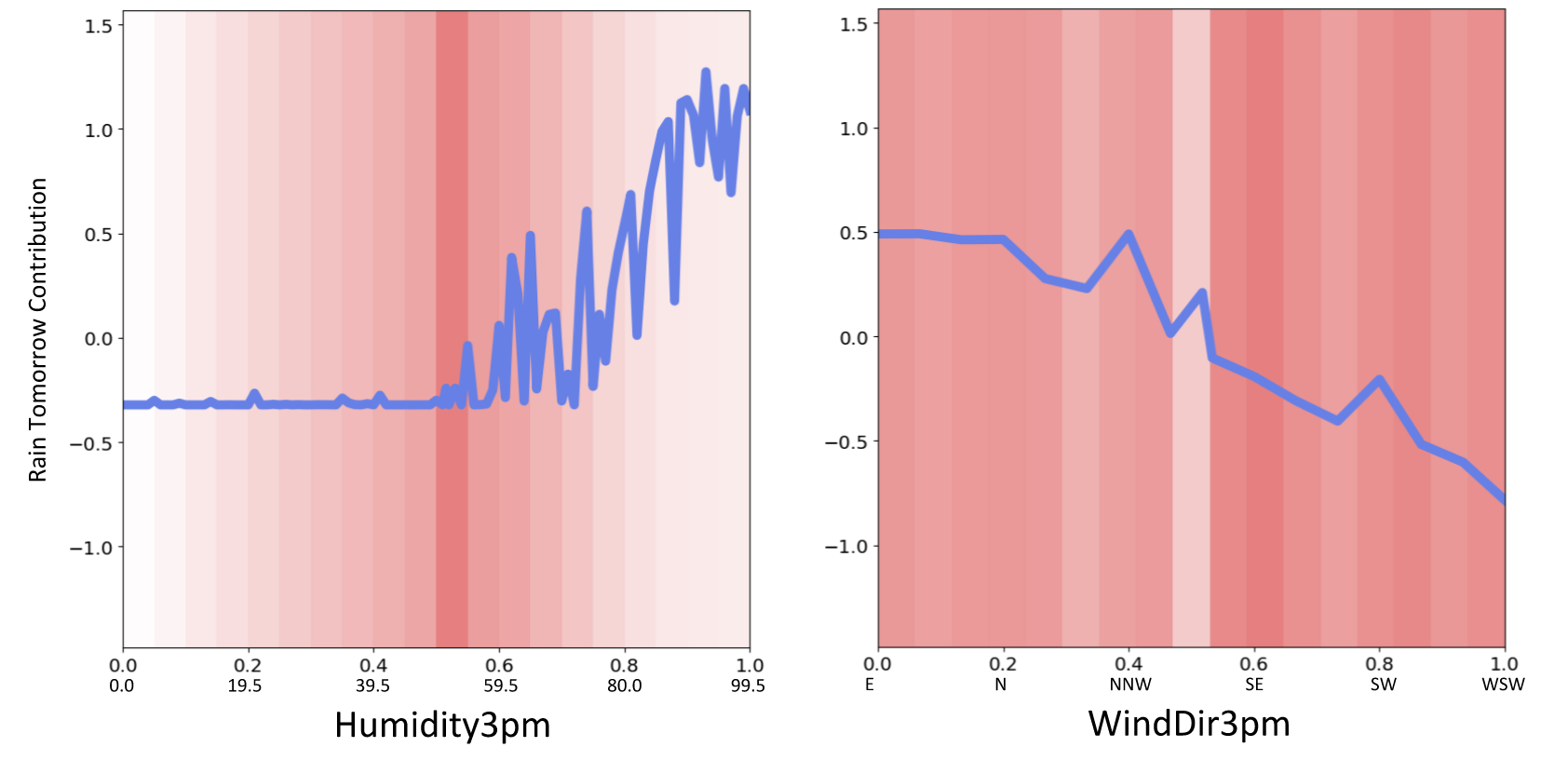} 
\label{fig: weather_nam}} 
\caption{NAM-style explanations for selected features learned by \mods. Network outputs are shown on the y-axis. Feature values are shown on the x-axis (normalised values above and actual values below). The blue line represents the learned shape function. Normalised data densities are shown using the red bars, the darker the red, the more data is available in that region.} 
\label{fig: explanation_results} 
\end{figure}

NAM generally performs worse in a classification task than in a regression task, achieving only 1.2\% and 24.4\% F1 score in the \textit{EEG} and \textit{Weather} data sets, respectively, as shown in \tablename~\ref{tab: comparison_results}. By incorporating \afs in its architecture, \qqmn overcomes NAM's limitation and produces the best results of all approaches on \textit{EEG}, while matching the performance of XGBoost with the best results on \textit{Weather}. This is discussed further in Section~\ref{sec: add_afs}.

\textbf{Explainability:} \qqmn is an interpretable model,  it produces explanatory plots showing the influence of selected features on model prediction. A selection of such plots is shown in Fig.~\ref{fig: explanation_results}, with further analyses of the explanations provided for each data set in Appendix C 
of the extended version of this paper \cite{Su2023}. As with NAM, the interpretability of \qqmn comes from visualising the shape of the plots for each modular network. Since each selected feature is learned independently, the shape functions show the exact contribution of a module (\ie feature) to a prediction. For example, it can be seen that there is a positive correlation between the quantity of carbon monoxide \textit{CO} and the predicted quantity of particulate matter $PM_{2.5}$.

Notice how in Fig.~\ref{fig: weather_nam}, the contribution of \textit{SO2} decreases linearly when the value of \textit{SO2} exceeds 320 $\mu g/mg^3$. This should warrant further investigation as the plot shows that there is insufficient data (with a lighter shade of red) when \textit{SO2} values exceed 100 $\mu g/m^3$. With these plots, the effect of \textit{CO} and \textit{SO2} on \textit{PM$_{2.5}$} can be further investigated, hopefully to produce a better understanding of the model and drive model intervention and improvement.

\section{Additional Experiments} \label{sec: add_exps}

\subsection{Effectiveness of \anb} \label{sec: add_anb}

In these additional experiments, we attempt to identify the impact of using an LSTM as the \rnn and that of using ANB as proposed in Section~\ref{sec: mod_main}. For the \rnn, we consider GRU and BLSTM as other options, while as an alternative to \anb we consider a standard linear layer and a variation of the exponential linear unit with inputs shifted by a bias (ExU), as used by NAM in \cite{Agarwal2021NAM}. For these experiments, all models are trained with the same hyper-parameters with 128 batch size, 0.002 learning rate, 128 hidden RNN units, 64 and 32 hidden units for the second and third linear layers in each module (\ie Layers 2 and 3 in Fig.~\ref{fig: architecture} \myfour , respectively), trained for 100 epochs with the Adam optimiser with early call-backs to avoid over-fitting. 

\begin{table}[!ht]
    \centering
    \caption{Test set loss comparison between different RNN networks and Weighted Linear Layers in the \textit{Air} data set. }
    \begin{tabular}{llc}
        \hline
        \\[-1em]
         Layer 1 & \rnn & Test Set Loss (Air Data set) $\downarrow$ \\
         \hline
         \\[-1em]
         \anb & LSTM & \textbf{0.0070}\\
          & GRU & 0.0072\\
          & BLSTM & 0.0216 \\
         Linears & LSTM  & 0.0216 \\
          & GRU & 0.0251\\
          & BLSTM & 0.0338\\
         ExU & LSTM & 0.0513 \\
          & GRU & 0.0591\\
          & BLSTM & 0.1334\\
         \hline
    \end{tabular}
    \label{tab: anb_loss}
\end{table}

Using the \textit{Air} data set, \tablename~\ref{tab: anb_loss} shows that the LSTM+\anb combination achieves the lowest loss. More importantly, any combination with ExU performed worse than its \textit{Linear} and \textit{\anb} counterparts. It can be seen, even before other data sets are considered, that \anb can improve the performance of \qqmn, with LSTM+\anb, GRU+\anb and BLSTM+\anb being the best performers. 

\subsection{Effectiveness of \afs} \label{sec: add_afs}

As observed in \tablename~\ref{tab: comparison_results}, NAM tends to perform worse in the classification tasks than in the regression tasks. Therefore, to further validate the effectiveness of the proposed \afs, we examine the effect of changing the number of features to be selected by  \afs. Using the \textit{Weather} classification data set, we compared test set accuracy when  $n=10$ (reported in \tablename~\ref{tab: comparison_results}) with $n=21$ (\ie including all available features). Optimal hyper-parameters already obtained for the \textit{Weather} data set are used, with both networks trained for 100 epochs on the Adam optimiser. The results show that the test set accuracy when $n=10$ is more than double the test set accuracy when all the features are used ($n=21$). 

\section{Conclusion and Future Work}

An interpretable modular neural network named \qqmn that uses attention for feature learning was introduced and evaluated in this paper. It has two main innovations: \afslong{} (\afs) and \anblong{} (\anb). In \afs, Multi-head Attention is modified to identify the most relevant features from the original input, while \anb acts as a bridge between \afs and each module of the interpretable network by weighting each selected feature based on its learned attention.

The goal of \qqmn was to improve predictive performance compared to NAM and yet retain interpretability. Experiments have shown that \qqmn not only outperforms NAM on all metrics (and its variant SPAM), but it can match the predictive performance of state-of-the-art non-interpretable models LSTM and XGBoost, outperforming the state-of-the-art in some cases. 

Related work AFS, MFS and TabNet share some similarities with the proposed \qqmn; however, \qqmn is distinct from them because attention weights are learned from a recurrent network in \qqmn and used for weighted feature selection. \qqmn also has flexibility using any recurrent neural networks. \qqmn was designed to be  closely related to architectures with modularity, such as additive models. A common assumption of such modules such as GAM and NAM is feature independence. This has been proved to be too strong an assumption in certain areas of application such as time series, as indicated by the poor predictive performance of such models in the experiments reported in this paper. \qqmn tackles this problem by using \afs for feature selection during training. Variations of NAM on the other hand, such as SPAM, seek to tackle the same problem by allowing interaction terms. This, however, may increase the complexity of the explanations provided for these models. As future work, we shall evaluate this and other trade-offs, such as the number of modules to use, in practice, since it is likely that the answers to these questions will depend on the characteristics of the specific application domain and quantitative expert evaluation.
  
In comparison with SPAM, a given application may require the use of higher-order feature interactions, not available in \qqmn. We shall investigate the possibility of the attention mechanism creating such features for implementation into a single network module. Finally, we shall continue to investigate the practical value of the explanations provided, how they may inform model intervention, and the interactions that may exist between attention-based explanations of \rnn{}s and interpretable modular neural networks. 

\section*{Acknowledgment}

This work was supported by the European Commission's Horizon 2020 research and innovation program under the SMART BEAR project (Grant agreement number: 8557172/H2020-SC1-FA-DTS-2018-2.) The authors would like to express their gratitude to the anonymous reviewers for their very helpful comments and to the authors of \cite{Christensen2021Oticon} for giving us access to the \emph{OtiReal} data set.



\bibliography{reference}

\clearpage
\appendix\input{appendix}


\end{document}

%% file: appendix.tex
\subsection{Optimal Hyper-parameters} \label{sec: optimal_hp}

The optimal hyper-parameters for each model for each dataset are listed below. Hyper-parameters are hyper-tuned using Ray-tune, specifically the Asynchronous Successive Halving Algorithm scheduler \cite{Li2018ASHA} with the Optuna search algorithm\footnote{\url{https://www.ray.io/}, \url{https://optuna.org/}} for better parallelism, on the validation set. 

\begin{table}[ht]
    \centering
    \caption{Optimal Hyper-Parameters}
    \begin{adjustbox}{width=\columnwidth}
    \begin{tabular}{llllll}
         \hline
         \\[-1em]
         & Hyper-parameters & OtiReal & Air & EGG & Weather \\
         \hline
         \\[-1em]
         \multirow{7}{*}{\qqmn} &
         Initial Learning Rate & 0.0130 & 0.0008 & 0.0021 & 0.0003 \\
         \\[-1em]
         & Dropout rate in the \rnn & 0.0318 & 0.0349 & 8.6209$e^5$ & 0.1992\\ 
         \\[-1em]
         & Dropout rate in the \mods & 0.1450 & 0.0142 & 0.0005 & 0.8378 \\
         \\[-1em]
         & Dropout rate in the final output & 0.0339 & 0.0513 & 0.0011 & 0.1851\\
         \\[-1em]
         & Batch Size & 256 & 64 & 512 & 64 \\
         \\[-1em]
         & Number of hidden unit in the \rnn & 256 & 64 & 64 & 1,026 \\ 
         \\[-1em]
         & Number of hidden units in the \mods & [256, 128] & [256, 128] & [32, 16] & [512, 256]\\
         \hline
         \multirow{4}{*}{NAM}
         \\[-1em]
         & Learning Rate & 0.0065 & 0.0122 & 0.0007 & 0.0002\\
         \\[-1em]
         & Dropout Rate & 0.4875 & 0.0405 & 0.0107 & 0.0218\\
         \\[-1em]
         & Batch Size & 64 & 512 & 256 & 256\\
         \\[-1em]
         & Hidden Sizes & [128, 64] & [512, 256] & [64, 32] & [256, 128]\\
         \hline
         \multirow{3}{*}{SPAM}
         \\[-1em]
         & Learning Rate & 0.268 & 0.1902 & 0.0009 & \\
         \\[-1em]
         & Dropout Rate & 0.1072 & 0.0265 & 0.0590 & \\
         \\[-1em]
         & Batch Size & 64 & 256 & 64 & \\
         \hline
         \multirow{4}{*}{LSTM}
         \\[-1em]
         & Learning Rate & 0.0260 & 0.0082 & 0.7330 & 0.0369\\
         \\[-1em]
         & Dropout rate & 0.1153 & 0.0567 & 0.2340 & 0.1285\\
         \\[-1em]
         & Batch Size & 256 & 64 & 128 & 64\\
         \\[-1em]
         & Number of hidden unit & 64 & 128 & 128 & 256\\
         \hline
         \multirow{4}{*}{XGBoost}
         \\[-1em]
         & Number of estimators & 1,000 & 200 & 50 & 1,000 \\
         \\[-1em]
         & Max Depths & 100 & 50 & 5 & 5\\
         \\[-1em]
         & Subsample & 0.2605 & 0.8069 & 0.3804 & 0.3191\\
         \\[-1em]
         & Learning Rate & 0.3370 & 0.7600 & 2.8467$e^5$ & 0.4725\\
         \hline
    \end{tabular}
    \end{adjustbox}
\end{table}

\subsection{Datasets Description} \label{sec: datasets_description}

\begin{table}[ht]
    \centering
    \caption{Datasets Description}
    \begin{adjustbox}{width=\columnwidth}
    \begin{tabular}{cccccc}
        \hline
        Name & OtiReal & Air & EEG & Weather \\ 
        \hline
        Source & \cite{Christensen2021Oticon} & \cite{Zhang2017Air} & \cite{Roesler2013EEG} & http://www.bom.gov.au/climate/data/ \\
        Instances & 24,736 & 420,768 & 14,892 & 137,910 \\
        Features & 16 & 16 & 14 & 21 \\
        Class Distribution & - & - & 1: 0.82 & 1: 0.26 \\
        Feature Type & Mixed & Mixed & Continuous & Mixed \\
        Top Features & 10 & 10 & 10 & 10 \\
        \hline
    \end{tabular}
    \end{adjustbox}
    \label{dataset_desc}
\end{table}

Real-world longitudinal and observational hearing aid (HAid) data are collected from patients signed up for the HearingFitness\textsuperscript{TM} feature with the Oticon ON\textsuperscript{TM} remote control app on the Oticon Open Hearing Aids (Oticon A/S, Smørum, Denmark). Christensen \etal \cite{Christensen2021Oticon} extracted a sampled data (\textit{OtiReal}) of 98 users between June-December 2019 that contains no personal information characterising the patients to preserve privacy. \textit{OtiReal} contains sound data concerning the acoustic environment detected by the HAids and logged by the connected smart-phones. These sound data include Sound Pressure Level (SPL), Modulation Index (MI), Signal-to-Noise Ratio (SNR), and Noise Floor (NF), all measured in a broadband frequency range of 0-10kHz in decibel units. HAid usage is calculated from the timestamps when the sound data are collected. The task is then to predict future daily HAid usage. In the Beijing Multi-Site Air-Quality (\textit{Air}) data set \cite{Zhang2017Air}, 7 years of meteorological data along with 4 years of $\text{PM}_{2.5}$ data of Beijing at 36 monitoring sites were collected. $\text{PM}_{2.5}$ refers to the fine particulate matter (PM) concentration in the air with aerodynamic diameter of less than 2.5 $\mu m$. The tasks is to predict future $\text{PM}_{2.5}$ values. 

In the EEG Eye State (\textit{EEG}) data set, continuous electroencephalogram (EEG) measurement are collected from the Emotive EEG Neuroheadset \cite{Roesler2013EEG}. The duration of the EEG measurement is 119 seconds, and eye state (\eg whether the eyes are closed or open) are detected by a camera during EEG measurement and added to the dataset manually later, where 1 indicates eye-closed and 0 indicates eye-open. The task is to classify eye state based on the EEG measurement. Another binary classification dataset is the Rain in Australia (\textit{Weather}) dataset, which contains 10 years of daily weather observations drawn from a number of weather stations in Australia. The task is to predict whether or not it will rain tomorrow. 

\subsection{Model Explanations}\label{sec: full_nam}

\subsubsection{Explanations for OtiReal Dataset}

\begin{figure}[!ht]
\centerline{\includegraphics[width=\linewidth]{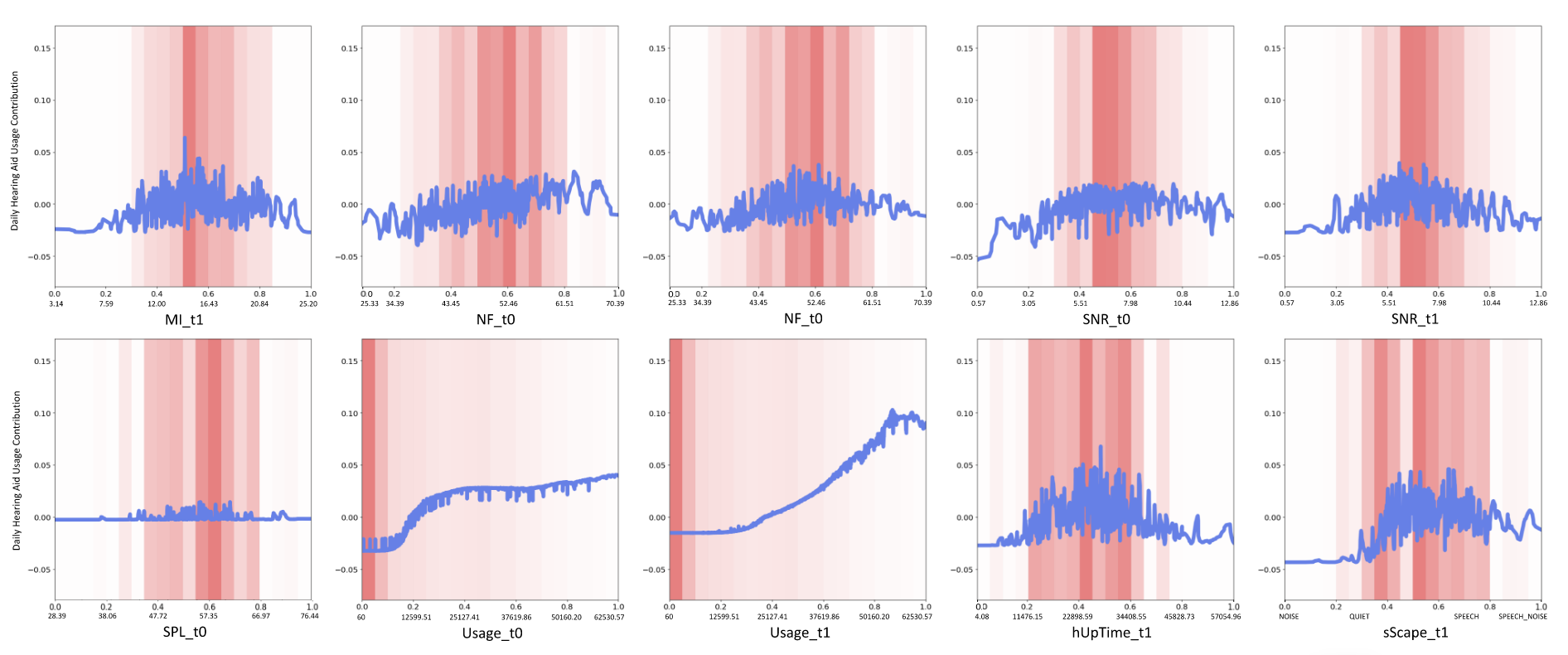}}
\caption{\textbf{OitReal} Graphs learned by \qqmn in predicting future hearing aid usage (regression) on the \textit{OtiReal} dataset. These plots show top 10 features selected by the \afs component in the \qqmn, where selected features with normalised and original values are on the x-axis, and daily future hearing aid usage prediction contribution are on the y-axis. In OtiReal, two timesteps of data are transformed and used as inputs, such that we are using data at $t_i$ and $t_{i+1}$ to predict hearing aid usage at $t_{i+2}$.} \label{fig: otireal_nam}
\end{figure}

The plots for \textit{MI}, \textit{NF}, and \textit{SNR} show that patients tend to use their HAids more in the future if the median values of these variables are sensed by their HAids. The lack of contribution of \textit{SPL} warrants further investigation with the audiologists as \textit{MI} and \textit{SNR} are derived from \textit{SPL}. The \textit{Soundscape} (sScape) classifies momentary sound environment into four categories by a proprietary HAid algorithm using \textit{MI}, \textit{SNR}, and \textit{SPL} values. Its plot shows that patients tend to use their HAids more if the soundscape is at the speech setting, whereas patients tend to use their HAids less with the noise or quiet setting. This could be an interesting observation to the audiologists as this might show that those patients who actively use their HAids to assist their hearing loss will tend to continuously use their HAids more in the future. The plot for \textit{Hearing Aid Up Time} (hUpTime) shows patients will be most likely to use their HAids in the future if their HAids have been activated for more than 23,000 seconds (or 6.3 hours). The plots for \textit{Usage} show that patients with higher daily HAid usage are more likely to have higher future daily HAid usage, and patients with lower daily HAid usage are more likely to have lower future daily HAid usage. These plots also show that the maximum contribution to future HAid usage of \textit{Usage$_{t0}$} and \textit{Usage$_{t1}$} is 0.05 and 0.1, respectively. This means that patients are more likely to have an even higher future daily HAid usage if they use their HAids for a sufficiently long period in a day continuously (for at least 2 days in this case). 

\subsubsection{Explanations for Air Dataset}

\begin{figure}[!h]
\centerline{\includegraphics[width=\linewidth]{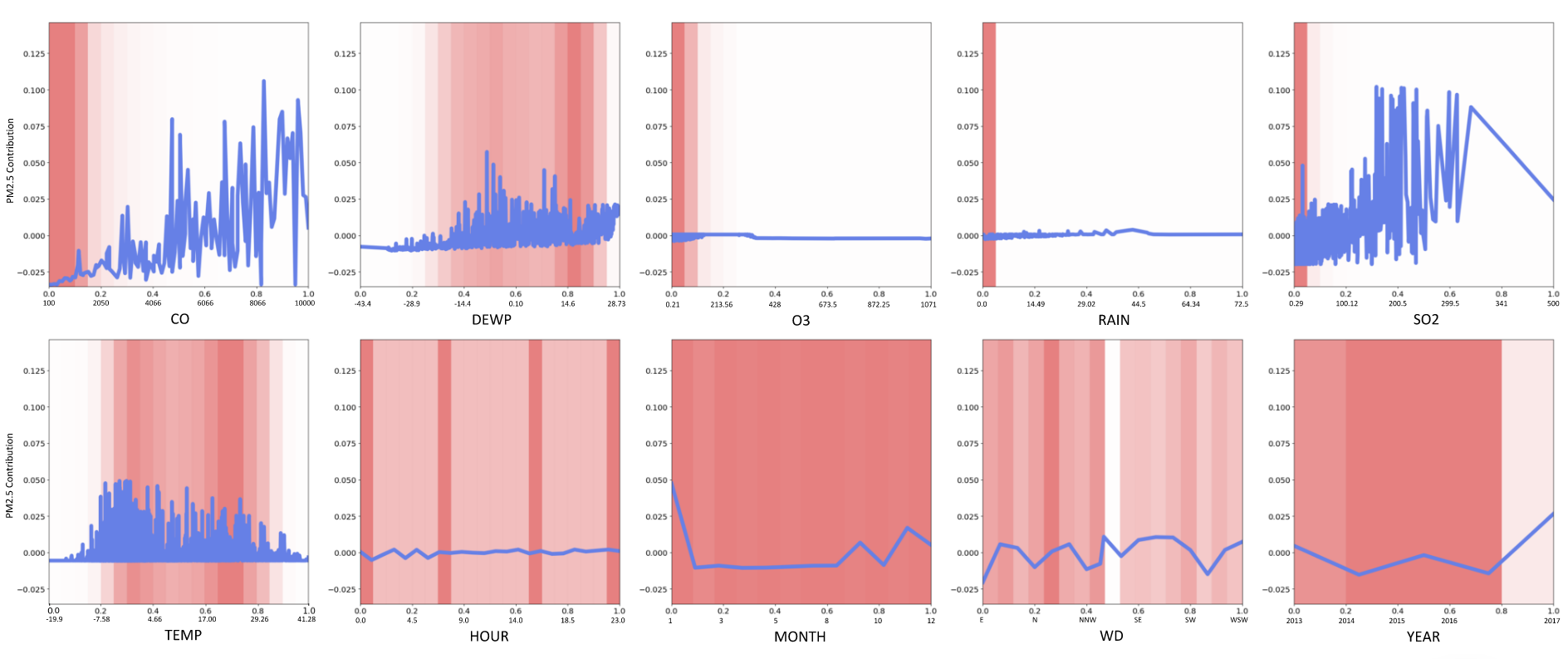}}
\caption{\textbf{Air} Graphs learned by \qqmn in predicting future PM$_{2.5}$ values  (regression) on the \textit{Air} dataset. These plots show top 10 features selected by the \afs component in the \qqmn, where selected features with normalised and original values are on the x-axis, and future PM$_{2.5}$ values prediction contribution are on the y-axis.} \label{fig: air_nam_full}
\end{figure}

The plot for \textit{Dew Point Temperature} (DEWP) shows that when \textit{DEWP} is at around 0 degree Celsius, this would lead to highest predicted future \textit{PM$_{2.5}$} values. \textit{O3 Concentration} (O3), \textit{Precipitation} (RAIN), \textit{HOUR} contribute almost nothing towards predicting future \textit{PM$_{2.5}$} values, and the contribution of \textit{O3} even becomes negative with high \textit{O3} values. The plot for \textit{Temperature} (TEMP) shows that when temperature is in the range of -8 and 28 degrees Celsius, future \textit{PM$_{2.5}$} values are predicted to be the highest, whereas when the temperature is above 28 and below -8 degree Celsius, this will lead to a low future \textit{PM$_{2.5}$} values. This observation is in line with many literature studying correlation between \textit{PM$_{2.5}$} and season such that there is a negative correlation in summer and autumn and positive correlation in spring and winter. The plot for \textit{MONTH} confirms this observation, such that future \textit{PM$_{2.5}$} values are predicted to be the highest in January, followed by November and September. The plot for \textit{Wind Direction} (WD) shows that future \textit{PM$_{2.5}$} values are predicted to be the lowest when there is a Eastern wind, and future \textit{PM$_{2.5}$} values are predicted to be the highest when the wind changes from a northerly to southerly wind. Lastly, the plot for \textit{YEAR} shows that more recent data (\ie data from 2016 onward) have more impact on the model in predicting higher future \textit{PM$_{2.5}$} values. 

\begin{figure}[!ht]
\centerline{\includegraphics[width=\linewidth]{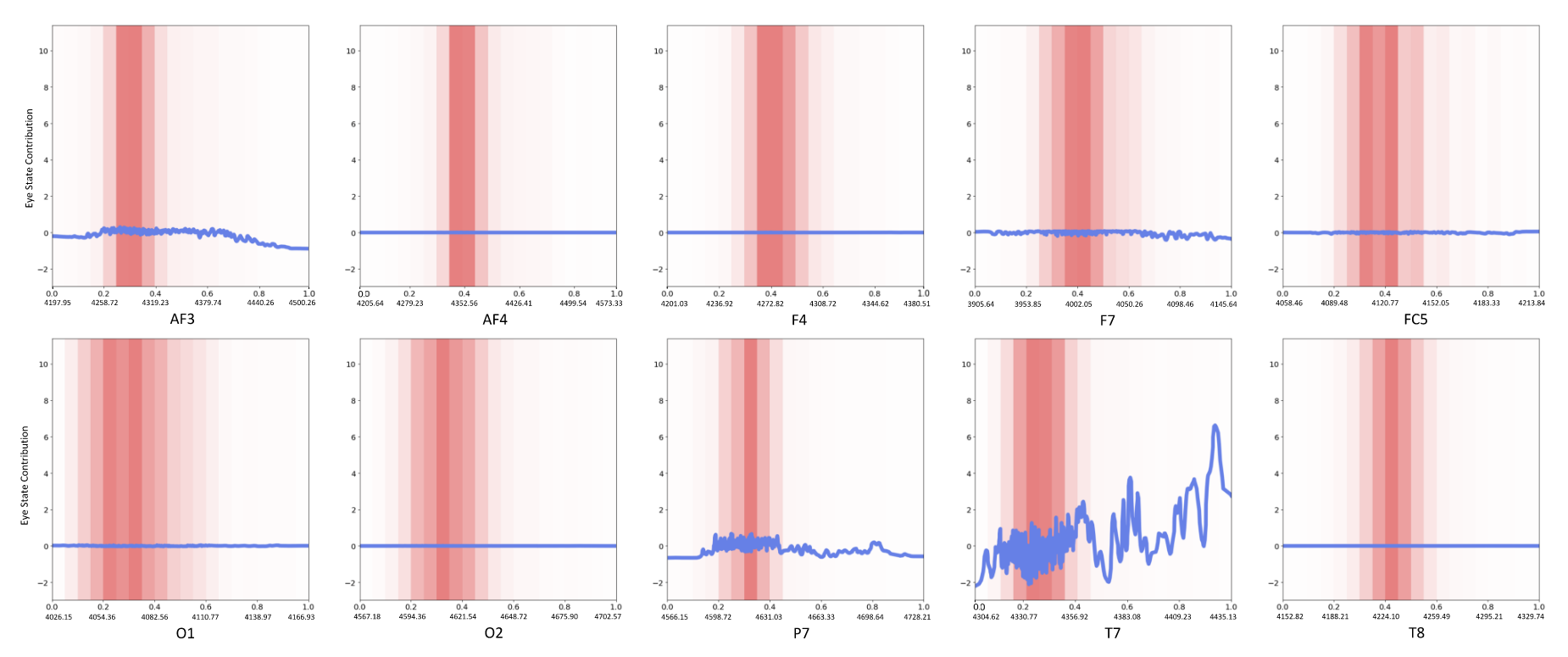}}
\caption{\textbf{EEG} Graphs learned by \qqmn in predicting eye state  (classification) on the \textit{EEG} dataset. These plots show top 10 features selected by the \afs component in the \qqmn, where selected features with normalised and original values are on the x-axis, and contribution towards the prediction are on the y-axis.} \label{fig: eeg_nam_full}
\end{figure}

\subsubsection{Explanations for EEG Dataset} \label{sec: eeg_exp}

Fig.~\ref{fig: eeg_nam_full} shows that \textit{AF4}, \textit{F4}, \textit{F7}, \textit{FC5}, \textit{O1}, \textit{O2}, and \textit{T8} have almost no contribution towards \qqmn in predicting eye state from the EEG, and \textit{AF3} and \textit{P7} have minimal contribution towards the model prediction. This means that electrodes placed at these regions do not contribute much to \qqmn in predicting the eye state. On the other hand, the electrode placed on the \textit{T7} region is the most important one in predicting the eye state. Its plot shows that higher \textit{T7} values means it is more likely that the eyes are open (\ie label 1), and lower \textit{T7} values means it is more likely that the eyes are closed (\ie label 0). Interestingly, \textit{T7} is a feature that is neither the most correlated to the outcome nor the one with the highest multicollinearity. Meaning that \qqmn managed to capture the non-linear relationship in the underlying data. 

\subsubsection{Explanations for Weather Dataset}

\begin{figure}[!ht]
\centerline{\includegraphics[width=\linewidth]{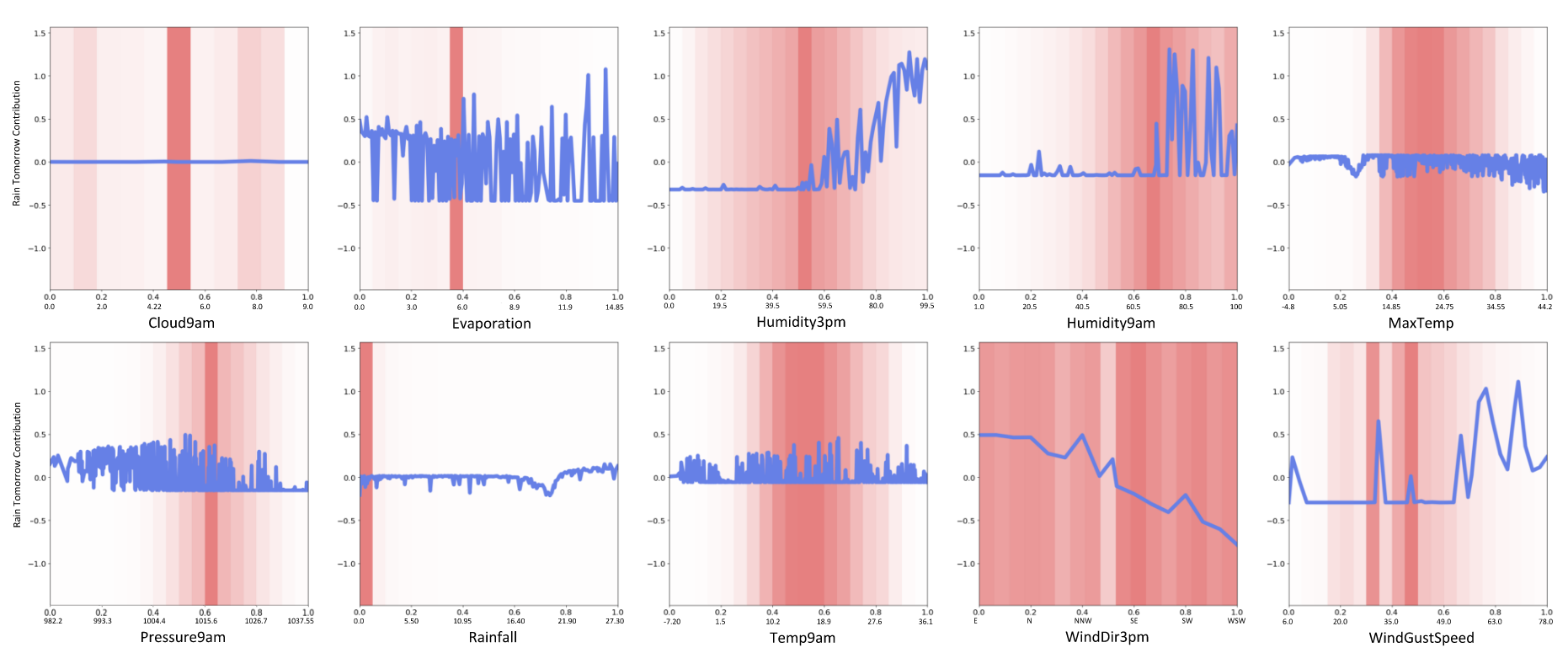}}
\caption{\textbf{Weather} Graphs learned by \qqmn in predicting whether it will rain tomorrow  (classification) on the \textit{Weather} dataset. These plots show top 10 features selected by the \afs component in the \qqmn, where selected features with normalised and original values are on the x-axis, and contribution towards the prediction are on the y-axis.} \label{fig: weather_nam_full}
\end{figure}

The plot for \textit{Cloud9am} shows that this feature contributed nothing in \qqmn predicting whether or not it will rain tomorrow. The relationship between evaporation and rain fall is rather complicated as it depends external factors including global climate phenomena, nevertheless, the shape plot for \textit{Evaporation} provides a detailed analysis such that when \textit{Evaporation} reaches 13mm, it is most likely lead to rain tomorrow. This complicated relationship can also be observed with \textit{Pressure9am} and \textit{Temp9am}, and how atmospheric pressure and temperature affects the chance of raining the next day also depends on other factors. The plot for \textit{Humidity9am} shows a drastic increase in the chance of raining tomorrow when the humidity is above 60\%. The plot for \textit{Maximum Temperature} (MaxTemp) shows that higher maximum temperature means there is a less chance of raining tomorrow. The plot for \textit{Rainfall} shows that, as expected, the more rainfall detected on the recorded day, more likely to rain tomorrow. There is a drop in chance of raining tomorrow when rainfall detected on the recorded day is above 17mm, and this might warrant a further investigation with the meteorologist. The plot for \textit{WindGustSpeed} shows that when the speed of the strongest wind is at around 70km/h, it is more likely to rain tomorrow.